\documentclass[letterpaper]{article} 

\usepackage{aaai2026}  
\usepackage{times}  
\usepackage{helvet}  
\usepackage{courier}  
\usepackage[hyphens]{url}  
\usepackage{graphicx} 
\usepackage{natbib}  
\usepackage{caption} 
\usepackage{amssymb,amsthm,amsmath,bm,mathtools,mathrsfs}
\usepackage{graphicx}
\usepackage{url}
\usepackage{centernot}
\usepackage{tabularx}
\usepackage{booktabs}
\usepackage[nameinlink]{cleveref}

\newcommand{\planningFont}[1]{#1}
\newcommand{\logicFont}[1]{\mathcal{#1}}
\newcommand{\graphFont}[1]{\mathbf{#1}}

\newcommand{\domain}{\planningFont{D}}
\newcommand{\predicates}{\planningFont{P}}
\newcommand{\schemata}{\planningFont{A}}

\newcommand{\ctwo}{$\logicFont{C}_2$}
\newcommand{\structure}{\logicFont{A}}
\newcommand{\langu}{\logicFont{L}}
\newcommand{\constants}{\logicFont{C}}
\newcommand{\structureFacts}{\logicFont{S}}

\newcommand{\nodes}{\graphFont{V}}
\newcommand{\edges}{\graphFont{E}}
\newcommand{\graph}{\graphFont{G}}

\newcommand{\goalNotAchieved}{\phi_{\mathrm{goalNotAchieved2}}}
\newcommand{\gnn}{\mathrm{GNN}}

\def\N{\mathbb{N}}

\def\e{\varepsilon}

\renewcommand{\phi}{\varphi}

\newcommand{\abs}[1]{| #1 |}

\newcommand{\gen}[1]{\left< #1 \right>}

\newcommand{\set}[1]{\left\{ #1 \right\}}

\newcommand{\lr}[1]{\left( #1 \right)}

\title{Relational GNNs Cannot Learn \ctwo{} Features for Planning}
\author{
    Dillon Z. Chen
}
\affiliations{
    LAAS-CNRS, University of Toulouse \\
    dchen@laas.fr
}

\nocopyright

\begin{document}

\maketitle


\begin{abstract}
Relational Graph Neural Networks (R-GNNs) are a GNN-based approach for learning value functions that can generalise to unseen problems from a given planning domain.
R-GNNs were theoretically motivated by the well known connection between the expressive power of GNNs and \ctwo{}, first-order logic with two variables and counting.
In the context of planning, \emph{\ctwo{} features} refer to the set of formulae in \ctwo{} with relations defined by the unary and binary predicates of a planning domain.
Some planning domains exhibit optimal value functions that can be decomposed as arithmetic expressions of \ctwo{} features.
We show that, contrary to empirical results, R-GNNs cannot learn value functions defined by \ctwo{} features.
We also identify prior GNN architectures for planning that may better learn value functions defined by \ctwo{} features.
\end{abstract}


\section{Introduction}
Graph neural networks (GNNs) are a popular deep learning architecture that can learn embeddings for graph data.
It is well known that the expressive power of standard GNNs is upper-bounded by the Weisfeiler-Leman (WL) algorithm~\cite{xu.etal.2019,morris.etal.2019}.
The expressive power of the WL algorithm and, hence GNNs, is in turn characterised by \ctwo{} logic~\cite{cai.etal.1989,barcelo.etal.2020,grohe.2021}, first-order logic with two variables and counting.
Inspired by this result, \citet{staahlberg.etal.2022} introduce Relational Graph Neural Networks (R-GNNs) for learning domain-dependent value functions for planning: value functions that are trained from a finite set of problems from a given classical planning domain that can be used in unseen problems from the same domain.
The key contributions of the paper include the characterisation of optimal value functions for multiple planning domains via \ctwo{} features, and corresponding experiments showing that R-GNNs can return optimal solutions with learned value functions on such domains.

However, it was never theoretically shown that R-GNNs can represent such value functions encoded via \ctwo{} features.
Indeed, we show in this paper that R-GNNs cannot learn such value functions using the counterexamples in \cite[Theorem 4.3]{chen.etal.2024} and \cite[Theorem 4.2]{chen.etal.2024a}.
The main reason for why this is the case is that the relations defined in R-GNNs do not correspond to the relations defined in \ctwo{} features, and thus the theory characterising the expressiveness of GNNs and \ctwo{} logic do not apply.
We instead show that the GNN architectures proposed by \citet{rivlin.etal.2020} and \cite{silver.etal.2021}, where objects are encoded as graph nodes and binary predicates as graph edges, exactly capture the \ctwo{} features described by \citet{staahlberg.etal.2022}.

The remainder of the paper is as follows.
\Cref{sec:prelim} defines notations of concepts relevant for results presented in the paper.
\Cref{sec:fail} highlights theoretical and empirical results stating that R-GNNs do not encapsulate \ctwo{} features.
\Cref{sec:good} identifies GNN architectures that exhibit a closer connection to \ctwo{} features.

\section{Background and Notation}\label{sec:prelim}
We begin by describing the classical planning instances represnted in lifted STRIPS.
We then introduce the concept of a relational language and relational structure in order to define \ctwo{} logic and \ctwo{} features for planning.

\paragraph{Planning Representation}
Following \citet{staahlberg.etal.2022}, a planning instance is a pair $P = \gen{D, I}$ where $D$ is a first-order planning domain and $I$ is the instance information.
A domain $D$ consists of a set of first-order predicates $\predicates$ and action schemata $\schemata$.
A predicate $p \in \predicates$ has parameters $x_1, \ldots, x_{n_p}$ for $n_p \in \N \cup \set{0}$, and $n_p$ denotes the arity of the predicate; e.g. a binary predicate $P$ has $n_p = 2$.
An action schema has preconditions and effects defined as formulae over atoms $p(x_1, \ldots, x_{n_p})$ where the $x_i$ are arguments of the schema.
The instance information $I$ consists of a finite set of object names $O = \set{o_1, \ldots, o_c}$, and an initial state $s_0$ and a goal condition $g$ which are sets of ground atoms: predicates in $\predicates$ whose parameters have been instantiated with object names in $O$; e.g. $p(o_3, o_2, o_4)$.
A state $s$ is a goal state if it contains all the goal condition atoms; i.e. $s \supseteq g$.

\paragraph{Relational Languages and Structures}
A relational language $\langu$ is a set of first-order predicates and their corresponding parameters and (nonnegative) arities.
A relational structure is a tuple $\structure = (\constants, \langu, \structureFacts)$ where $\constants$ is a set of constants, $\langu$ is a relational language, and $\structureFacts$ a set of ground atoms: predicates in $\langu$ whose parameters have been instantiated with constants from $\constants$.

A graph is an example of a relational structure with relational language induced from the edge relations of graphs.
Specifically, a graph $\graph$ defined as a tuple of nodes $\nodes$ and edges $\edges \subseteq {{\nodes}\choose{2}}$ induces the relational structure $\structure_\graph = (\nodes, \set{E}, \set{E(u, v) \mid (u, v) \in \edges})$, where $E$ is a binary predicate corresponding to graph edges.

\paragraph{\ctwo{} Logic}
\ctwo{} logic refers to first-order logic with 2 variables and counting over a given relational language.
We recite two example formulae by~\citet{cai.etal.1989} for the relational language for graphs and refer to the same article for a formal presentation of \ctwo{} logic.

The first such example does not use counting:
$
    \phi \equiv \forall x \exists y (E(x, y) \wedge \exists x [\neg E(x, y)]).
$
The formula $\phi$ states that every vertex is adjacent to some other vertex that itself is not adjacent to every vertex in the graph.
Note that variables may be requantified, as in the case of $x$ here.
The second such example involves counting:
$
    \psi \equiv (\exists^{\geq 17}x)(\exists^{\geq 5}y)E(x, y).
$
The formula $\psi$ states that there are at least 17 vertices of degree at least 5.
In general, one can also specify exact values with the expression
$
    (\exists^{=i} x) \phi(x) \equiv (\exists^{\geq i}x)\phi(x) \wedge \neg(\exists^{\geq i+1}x) \phi(x)
$
which states that there are exactly $i$ constants satisfying the formula $x$ (at least $i$ constants but not more than $i+1$ constants).

\paragraph{\ctwo{} Features for Planning}
Given a planning domain $\domain$, \citet{staahlberg.etal.2022} refer to \ctwo{} features as formulae in \ctwo{} logic with the language $\predicates \cup \predicates_G$ where $\predicates$ are the predicates of $\domain$ and $\predicates_G := \set{p_G \mid p \in \predicates}$ are duplicate predicates for encoding goal information.
A states in a given planning instance also induces a relational structure.
This is done by extending the state with a copy of the goal condition from their corresponding instance $s_G := \set{p_G(o_1, \ldots, o_{n_p}) \mid p(o_1, \ldots, o_n) \in g}$.
Then the explicit relational structure for a given state is given by
\begin{align}
    \structure_s = (O, \predicates \cup \predicates_G, s \cup s_G)
    \label{eqn:planning-structure}
\end{align}
where $O$ is the set of objects in the corresponding instance.
An example \ctwo{} feature for planning consists of checking if the goal has not been achieved for planning domains with predicates whose predicates have at most arity 2:
\newcommand{\llll}{\biggl(}
\newcommand{\rrr}{\biggr)}
\begin{align*}
     & \goalNotAchieved \equiv                                                                           \\
     & \qquad
    \llll\bigvee_{p(x_1,x_2) \in \predicates} \exists x \exists y (p_G(x,y) \wedge \neg p(x,y))\rrr \vee \\
     & \qquad\qquad
    \llll\bigvee_{p(x_1) \in \predicates} \exists x (p_G(x) \wedge \neg p(x))\rrr \vee                   \\
     & \qquad\qquad\qquad
    \llll\bigvee_{p() \in \predicates} (p_G() \wedge \neg p())\rrr
\end{align*}

The formula $\goalNotAchieved$ states that the goal is not achieved when there is at least one goal atom with predicate at most 2 that is not achieved in the state.
Note that $\goalNotAchieved$ does not make use of counting.
Indeed, a special case of $\goalNotAchieved$ is presented in \cite[page 3]{staahlberg.etal.2022}.
More specifically, one can define the formula $\alpha \equiv \exists x \exists y \lr{\planningFont{On}_G(x, y) \wedge \neg \planningFont(x, y)}$ for detecting whether a single Blocksworld goal atom has been achieved, and similarly for other planning domains.

\section{R-GNNs $\not=$ \ctwo{}}\label{sec:fail}
The proof for why R-GNNs do not encapsulate \ctwo{} features for planning is straightforward and involves showing that it cannot distinguish pairs of planning problems with different $\goalNotAchieved$ evaluations.
The counterexample is inspired by \cite[Theorem 4.3]{chen.etal.2024} and explicitly presented in \cite[Theorem 4.2]{chen.etal.2024a} which we describe as follows.

\paragraph{Counterexample}
Consider the planning domain $\domain$ with a single binary predicate $\predicates = \set{q(x, y)}$ and single action schema $\schemata = \set{s(x, y)}$.
The action schema $s(x, y)$ has no preconditions and its effect adds the atom $q(x, y)$.
Then consider two different planning instances with information in the following table.
\begin{table}[h!]
    \small
    \begin{tabularx}{\columnwidth}{l l l l}
        \toprule
        Instance & Objects        & Initial State          & Goal Condition             \\
        \midrule
        $I_1$    & {$\set{a, b}$} & $\set{q(a,a), q(b,b)}$ & {$\set{q(a, b), q(b, a)}$} \\
        $I_2$    & {$\set{a, b}$} & $\set{q(a,b), q(b,a)}$ & {$\set{q(a, b), q(b, a)}$} \\
        \bottomrule
    \end{tabularx}
\end{table}

Both planning instances $I_1$ and $I_2$ have the same objects and goal condition but only differ in their state.
Note that the optimal solution for $I_1$ has 2 actions ($s(a,b)$ and $s(b,a)$) while $I_2$'s state is already in the goal condition.
Thus the evaluation of $\goalNotAchieved$ is true on $I_1$ and is false on $I_2$.
In other words, if R-GNNs indeed capture \ctwo{} features, they should \emph{distinguish} $I_1$ and $I_2$; i.e. R-GNNs should return different outputs for $I_1$ and $I_2$ for some set of hyperparameters and weights.

However, \cite[Theorem 4.2]{chen.etal.2024a} shows that R-GNNs cannot distinguish $I_1$ and $I_2$, which implies that R-GNNs cannot learn nor capture \ctwo{} features.

\subsection*{Empirical Tests}
We also empirically show that the counterexample holds true.
We perform two different setups commonly performed in the GNN research community for empirically testing the distinguishing power of GNN architectures~\cite{kriege.etal.2020,abboud.etal.2021,balcilar.etal.2021,feng.etal.2022,zhao.etal.2022a,wang.etal.2023,bouritsas.etal.2023}.
Both setups are not entirely conclusive due to numerical precision errors, but are very likely to be representative of the expressive power of a GNN on specific inputs.
Given a GNN architecture $\gnn^\theta$ with hyperparameters and weights $\theta$, and a pair of inputs $I_1$ and $I_2$, denote $O^{\theta}_1 = \gnn_\theta(I_1)$ and $O^{\theta}_2 = \gnn_\theta(I_2)$ to be the GNN's outputs.

\def\maxx{\max(\abs{O^{\theta_i}_1}, \abs{O^{\theta_i}_2})}
\def\diff{\abs{O^{\theta_i}_1 - O^{\theta_i}_2}}
\paragraph{Distinguishibility Test by Random Initialisation}
The first experiment we perform involves randomly initialising $n$ different hyperparameter and weight configurations for the GNN to get $\theta_1, \ldots, \theta_n$.
Then if it holds that
\begin{align*}
    {\diff}/{\maxx} > \e
\end{align*}
for some small $\e > 0$ and some $\theta_i$, it is likely that $\gnn$ can distinguish the pair of inputs.
The $\e$ is nonzero to account for numerical precision errors from neural network predictions, and represents the percentage of the difference $\diff$ relative to the maximum $\maxx$.
Otherwise, it is likely that $\gnn$ cannot distinguish the pair and the likelihood increases with larger $n$ and smaller $\e$.

We perform this setup using publicly available code\footnote{Accessed from \url{https://github.com/simon-stahlberg/relational-neural-network-python} under the commit \texttt{dd70a46}} for R-GNN.
We fix the hyperparameters to be \texttt{64} for the embedding size; \texttt{sum} to be the aggregation function, as the sum function is maximally expressive for GNNs~\cite{xu.etal.2019}; and \texttt{30} to be the number of layers.

\begin{figure}[t]
    \centering
    \includegraphics[width=\columnwidth]{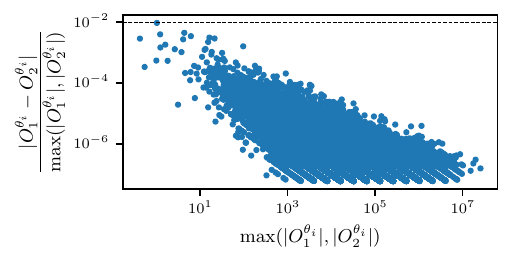}
    \caption{Relative difference of R-GNN predictions on $I_1$ and $I_2$ vs. the maximum of the predictions. The dotted horizontal line represents $y=0.01$.}
    \label{fig:random}
\end{figure}

Then we randomly initialise weights for R-GNN $n=100,000$ times and set $\e = 0.01$ (1\%) to account for numerical instability.
\Cref{fig:random} illustrates the results.
Over all $n=100,000$ trials, there is a less than 1\% difference between the predictions made on the input tasks relative to the maximum of the predictions.

\paragraph{Distinguishibility Test by Training}
Recall that the counterexample has two instances $I_1$ and $I_2$ with optimal plan lengths given by 2 and 0 respectively.
If R-GNN is able to distinguish between the two instances, then it should be able to learn their optimal plan lengths if trained on them directly.
We train the R-GNN on the dataset consisting of only $I_1$ and $I_2$ and their corresponding ground truth values 2 and 0.
The same hyperparamter configuration of R-GNN in the previous experiment is used, alongside the default optimiser provided in the code (Adam with a learning rate of $0.0002$).

The loss never drops below 1, the average of the ground truth values, across all 10000 training steps.
This is the loss that R-GNN can best achieve if it cannot distinguish the inputs.
This observation holds true for different tested hyperparameters for R-GNN, such as the default R-GNN hyperparameters provided in the code.

These two experiments imply that it is highly likely that R-GNN is not able to distinguish the two instances, supporting the result in \cite[Theorem 4.2]{chen.etal.2024a}.
More precisely, R-GNNs cannot learn value functions defined with \ctwo{} features for planning.

\section{PLOI Graphs + GNN $\simeq$ \ctwo{}}\label{sec:good}
We conclude this paper by identifying graph encodings of planning instances on which GNNs are better able to learn \ctwo{} features.
We identify them as PLOI graphs, named\footnote{Although the graph was first introduced by \citet{rivlin.etal.2020} in the context of learning, they do not introduce a descriptive name of their approach.} after the work by \citet{silver.etal.2021}.
We begin by introducing a more expressive notion of a graph under the relational structure framework, which includes node features and edge labels.
A graph with node features and edge labels is a relational structure $(\nodes, \set{E_1, \ldots, E_l, F_1, \ldots, F_n}, \structureFacts)$ where the $E_i$ are binary relations representing edge labels and the $F_j$ are unary relations representing node features.
Note that it is possible for a pair of nodes to be instantiated in two different relations $E_i$ and $E_j$, and similarly for singleton nodes in different unary relations $F_i$ and $F_j$.
Furthermore, a graph presented in this way can be transformed into a standard GNN input with node and edge features represented as one-hot encodings that encode whether the relations $F_j$ and $E_i$ hold respectively, and vice versa for GNN inputs with binary node and edge features as is the case for PLOI.

\paragraph{PLOI Graphs}
Let $s$ be a planning state, and $G$ be the goal condition and $O$ the objects from the corresponding planning instance, and $\predicates$ the set of predicates from the corresponding planning domain.
We assume that all predicates are at most binary\footnote{Higher arity predicates can be compiled down to binary predicates but the compilation loses information~\cite[Theorem 4.4]{chen.etal.2024a}} due to the two variable limitation of \ctwo{}.

Then one defines the PLOI graph with node features and edge labels as the relational structure
\begin{align}
    \structure_{\mathrm{PLOI}} = (
    O,
    \predicates \cup \predicates_G \cup \overleftarrow{\predicates} \cup \overleftarrow{\predicates_G},
    s \cup s_G \cup \overleftarrow{s} \cup \overleftarrow{s_G}
    )
    \label{eqn:ploi}
\end{align}
where
\begin{itemize}
    \item $\predicates_G$ and $s_G$ are the same as in \Cref{eqn:planning-structure},
    \item $\overleftarrow{\predicates} := \set{\overleftarrow{p} \mid p \in \predicates, \text{$p$ is binary}}$ and similarly for $\overleftarrow{\predicates_G}$, and
    \item $\overleftarrow{s} := \set{\overleftarrow{p}(o_2, o_1) \mid p(o_1, o_2) \in \predicates}$ and similarly for $\overleftarrow{s_G}$.
\end{itemize}

Notably, the PLOI relational structure defined in \Cref{eqn:ploi} subsumes the relational structure for planning defined in \Cref{eqn:planning-structure}.
Now recall that \ctwo{} features are defined from the planning structure.
Furthermore, GNNs can encapsulate \ctwo{} logic with unary and binary relations defined from the graphs they operate on~\cite{barcelo.etal.2020,grohe.2021}.
Thus, it is more sensible to view GNNs operating on the PLOI graph as better able to capture \ctwo{} features for planning than R-GNNs.

\section{Conclusion}
We highlighted theoretically and empirically that Relational GNNs for planning (R-GNNs)~\cite{staahlberg.etal.2022} do not correspond with \ctwo{} features for planning, \ctwo{} logic where relations are defined by unary and binary planning domain predicates.
In light of this, we identify that a different GNN architecture, PLOI~\cite{rivlin.etal.2020,silver.etal.2021}, that better approximates \ctwo{} features for planning.

\bibliography{relnn}

\begin{thebibliography}{17}
\providecommand{\natexlab}[1]{#1}

\bibitem[{Abboud et~al.(2021)Abboud, Ceylan, Grohe, and
  Lukasiewicz}]{abboud.etal.2021}
Abboud, R.; Ceylan, {\.I}.~{\.I}.; Grohe, M.; and Lukasiewicz, T. 2021.
\newblock The Surprising Power of Graph Neural Networks with Random Node
  Initialization.
\newblock In \emph{{IJCAI}}.

\bibitem[{Balcilar et~al.(2021)Balcilar, H{\'{e}}roux, Ga{\"{u}}z{\`{e}}re,
  Vasseur, Adam, and Honeine}]{balcilar.etal.2021}
Balcilar, M.; H{\'{e}}roux, P.; Ga{\"{u}}z{\`{e}}re, B.; Vasseur, P.; Adam, S.;
  and Honeine, P. 2021.
\newblock Breaking the Limits of Message Passing Graph Neural Networks.
\newblock In \emph{{ICML}}.

\bibitem[{Barcel{\'{o}} et~al.(2020)Barcel{\'{o}}, Kostylev, Monet,
  P{\'{e}}rez, Reutter, and Silva}]{barcelo.etal.2020}
Barcel{\'{o}}, P.; Kostylev, E.~V.; Monet, M.; P{\'{e}}rez, J.; Reutter, J.~L.;
  and Silva, J.~P. 2020.
\newblock The Logical Expressiveness of Graph Neural Networks.
\newblock In \emph{{ICLR}}.

\bibitem[{Bouritsas et~al.(2023)Bouritsas, Frasca, Zafeiriou, and
  Bronstein}]{bouritsas.etal.2023}
Bouritsas, G.; Frasca, F.; Zafeiriou, S.; and Bronstein, M.~M. 2023.
\newblock Improving Graph Neural Network Expressivity via Subgraph Isomorphism
  Counting.
\newblock \emph{{IEEE} Trans. Pattern Anal. Mach. Intell.}, 45(1).

\bibitem[{Cai, F{\"{u}}rer, and Immerman(1989)}]{cai.etal.1989}
Cai, J.; F{\"{u}}rer, M.; and Immerman, N. 1989.
\newblock An Optimal Lower Bound on the Number of Variables for Graph
  Identification.
\newblock In \emph{{FOCS}}.

\bibitem[{Chen, Thi{\'{e}}baux, and Trevizan(2024)}]{chen.etal.2024}
Chen, D.~Z.; Thi{\'{e}}baux, S.; and Trevizan, F. 2024.
\newblock Learning Domain-Independent Heuristics for Grounded and Lifted
  Planning.
\newblock In \emph{AAAI}.

\bibitem[{Chen, Trevizan, and Thi{\'{e}}baux(2024)}]{chen.etal.2024a}
Chen, D.~Z.; Trevizan, F.; and Thi{\'{e}}baux, S. 2024.
\newblock Return to Tradition: Learning Reliable Heuristics with Classical
  Machine Learning.
\newblock In \emph{ICAPS}.

\bibitem[{Feng et~al.(2022)Feng, Chen, Li, Sarkar, and Zhang}]{feng.etal.2022}
Feng, J.; Chen, Y.; Li, F.; Sarkar, A.; and Zhang, M. 2022.
\newblock How Powerful are K-hop Message Passing Graph Neural Networks.
\newblock In \emph{NeurIPS}.

\bibitem[{Grohe(2021)}]{grohe.2021}
Grohe, M. 2021.
\newblock The Logic of Graph Neural Networks.
\newblock In \emph{{LICS}}.

\bibitem[{Kriege, Johansson, and Morris(2020)}]{kriege.etal.2020}
Kriege, N.~M.; Johansson, F.~D.; and Morris, C. 2020.
\newblock A survey on graph kernels.
\newblock \emph{Appl. Netw. Sci.}, 5(1): 6.

\bibitem[{Morris et~al.(2019)Morris, Ritzert, Fey, Hamilton, Lenssen, Rattan,
  and Grohe}]{morris.etal.2019}
Morris, C.; Ritzert, M.; Fey, M.; Hamilton, W.~L.; Lenssen, J.~E.; Rattan, G.;
  and Grohe, M. 2019.
\newblock Weisfeiler and Leman Go Neural: Higher-Order Graph Neural Networks.
\newblock In \emph{{AAAI}}.

\bibitem[{Rivlin, Hazan, and Karpas(2020)}]{rivlin.etal.2020}
Rivlin, O.; Hazan, T.; and Karpas, E. 2020.
\newblock Generalized Planning With Deep Reinforcement Learning.
\newblock \emph{CoRR}, abs/2005.02305.

\bibitem[{Silver et~al.(2021)Silver, Chitnis, Curtis, Tenenbaum,
  Lozano{-}P{\'{e}}rez, and Kaelbling}]{silver.etal.2021}
Silver, T.; Chitnis, R.; Curtis, A.; Tenenbaum, J.~B.; Lozano{-}P{\'{e}}rez,
  T.; and Kaelbling, L.~P. 2021.
\newblock Planning with Learned Object Importance in Large Problem Instances
  using Graph Neural Networks.
\newblock In \emph{{AAAI}}.

\bibitem[{St{\aa}hlberg, Bonet, and Geffner(2022)}]{staahlberg.etal.2022}
St{\aa}hlberg, S.; Bonet, B.; and Geffner, H. 2022.
\newblock Learning General Optimal Policies with Graph Neural Networks:
  Expressive Power, Transparency, and Limits.
\newblock In \emph{ICAPS}.

\bibitem[{Wang et~al.(2023)Wang, Chen, Wijesinghe, Li, and
  Farhan}]{wang.etal.2023}
Wang, Q.; Chen, D.~Z.; Wijesinghe, A.; Li, S.; and Farhan, M. 2023.
\newblock {$\mathcal{N}$}-WL: {A} New Hierarchy of Expressivity for Graph
  Neural Networks.
\newblock In \emph{{ICLR}}.

\bibitem[{Xu et~al.(2019)Xu, Hu, Leskovec, and Jegelka}]{xu.etal.2019}
Xu, K.; Hu, W.; Leskovec, J.; and Jegelka, S. 2019.
\newblock How Powerful are Graph Neural Networks?
\newblock In \emph{{ICLR}}.

\bibitem[{Zhao et~al.(2022)Zhao, Jin, Akoglu, and Shah}]{zhao.etal.2022a}
Zhao, L.; Jin, W.; Akoglu, L.; and Shah, N. 2022.
\newblock From Stars to Subgraphs: Uplifting Any {GNN} with Local Structure
  Awareness.
\newblock In \emph{{ICLR}}.

\end{thebibliography}

\end{document}